\documentclass[conference, 11.6pt, onecolumn]{IEEEtran}

\usepackage{times}
\usepackage{latexsym} 
\usepackage{graphicx}
\usepackage[table]{xcolor}
\usepackage{rotating}
\usepackage{lipsum}
\usepackage[T1]{fontenc}
\usepackage[utf8]{inputenc}
\definecolor{dgreen}{RGB}{24,70,50}
\usepackage{hyperref}

\title{Accelerating Neural Network Training: A Brief Review}

\author{\IEEEauthorblockN{ Sahil Nokhwal}
\IEEEauthorblockA{\textit{Computer Science Dept.} \\
\textit{University of Memphis} \\
Memphis, USA \\
nokhwal.official@gmail.com}
\and

\IEEEauthorblockN{Priyanka Chilakalapudi}
\IEEEauthorblockA{\textit{ Computer Science Dept.} \\
\textit{University of Memphis}\\
Memphis, USA \\
pchlklpd@memphis.edu}

\and
\IEEEauthorblockN{Preeti Donekal}
\IEEEauthorblockA{\textit{Data Science Dept.} \\
\textit{University of Memphis}\\
Memphis, USA \\
pdonekal@memphis.edu}

\and
\IEEEauthorblockN{Suman Nokhwal}
\IEEEauthorblockA{\textit{ Staff Engineer } \\
\textit{Intercontinental Exchange, Inc.}\\
Pleasanton, California, USA \\
suman148@gmail.com}

\and
\IEEEauthorblockN{Saurabh Pahune}
\IEEEauthorblockA{\textit{Software Engineering} \\
\textit{Cardinal Health}\\
Dublin, Ohio, USA \\
saurabh.pahune@cardinalhealth.com} %saurabh.pahune   saurabhpahune214@gmail.com

\and
\IEEEauthorblockN{Ankit Chaudhary}
\IEEEauthorblockA{\textit{Associate Professor} \\
\textit{Jawaharlal Nehru University}\\
New Delhi, Delhi, India \\
dr.ankit@ieee.org}
}

\begin{document}
\maketitle
\thispagestyle{plain}
\pagestyle{plain}

\begin{abstract}
The process of training a deep neural network is characterized by significant time requirements and associated costs. Although researchers have made considerable progress in this area, further work is still required due to resource constraints. This study examines innovative approaches to expedite the training process of deep neural networks (DNN), with specific emphasis on three state-of-the-art models such as ResNet50, Vision Transformer (ViT), and EfficientNet. The research utilizes sophisticated methodologies, including Gradient Accumulation (GA), Automatic Mixed Precision (AMP), and Pin Memory (PM), in order to optimize performance and accelerate the training procedure.

The study examines the effects of these methodologies on the DNN models discussed earlier, assessing their efficacy with regard to training rate and computational efficacy. The study showcases the efficacy of including GA as a strategic approach, resulting in a noteworthy decrease in the duration required for training. This enables the models to converge at a faster pace. The utilization of AMP enhances the speed of computations by taking advantage of the advantages offered by lower precision arithmetic while maintaining the correctness of the model.

Furthermore, this study investigates the application of Pin Memory as a strategy to enhance the efficiency of data transmission between the central processing unit and the graphics processing unit, thereby offering a promising opportunity for enhancing overall performance. The experimental findings demonstrate that the combination of these sophisticated methodologies significantly accelerates the training of DNNs, offering vital insights for experts seeking to improve the effectiveness of deep learning processes.
\end{abstract}


\maketitle

\section{Introduction}
The task of image classification can be addressed by the application of many machine-learning methodologies. However, contemporary research has predominantly emphasized the utilization of deep-learning neural network architectures, including ResNet50 \cite{he2016deep}, Vision Transformer \cite{dosovitskiy2020image}, and EfficientNet \cite{tan2019efficientnet}. Despite the notable progress made by the computer vision and deep learning research communities in the field of image classification, there remains a dearth of efforts directed toward enhancing training speed, particularly when dealing with extensive image datasets. 

This study aims to demonstrate the relative training speed of the aforementioned model in the multi-class image classification domain. Additionally, the study aims to identify the model that surpasses other models in terms of accuracy, F1 score, and execution time, when evaluated against the selected datasets. 

When training a neural network on a solitary GPU, the "CUDA out-of-memory" problem is a frequent hardware constraint that must be considered. In addition, the training of a deep neural network is a laborious process that hinders the progress of scientific inquiry. 

It is crucial to acknowledge that although the present inquiry has concentrated on evaluating the performance strategies on three particular DNN models, these methods could be equally applicable to other DNN models. We have made our code available at \url{https://github.com/SahilNokhwal/annt/}. \\

The following are our main contributions:
\begin{enumerate}
    \item In this study, we suggest the use of three strategies, namely Gradient Accumulation, Automatic Mixed Precision, and Pin Memory, as a means to enhance the efficiency of neural network training. 
    \item A comparative analysis has been done to evaluate the efficacy of different performance optimization approaches on a range of DNN models.
    \item We have made our research source code available for further research in this area.
\end{enumerate}

% \begin{figure}[!htb]
% \centering
% \includegraphics[width=0.4\textwidth]{resources/resnet_arch.png}
% \caption{\label{fig:resnet50_arch}ResNet50 architecture\cite{arch_resnet}}
% \end{figure}

% \begin{figure}[!htb]
% \centering
% \includegraphics[width=0.4\textwidth]{resources/vit_arch.png}
% \caption{\label{fig:vit_arch}ViT architecture\cite{dosovitskiy2020image}}
% \end{figure}

% \begin{figure}[!htb]
% \centering
% \includegraphics[width=0.4\textwidth]{resources/efficientnet_arch.png}
% \caption{\label{fig:efficientnet_arch}EfficientNet architecture\cite{tan2019efficientnet}}
% \end{figure}

\section{Related Work}
 The SWATS approach, as suggested in \cite{keskar2017improving}, involves the substitution of the Adam optimizer \cite{kingma2014adam} with the SGD optimizer under certain circumstances. The suggested methodology introduces little additional computational burden, hence avoiding an increase in the total amount of hyperparameters. The researchers in \cite{nagpal2019performance} conducted a comprehensive assessment of the performance of CNNs in the context of face anti-spoofing. In \cite{mu2020history} introduced a novel auto-tuning methodology that aims to enhance the design of GPU-based operators. The authors of the study also demonstrated the effectiveness of their framework by achieving superior outcomes compared to the cutting-edge auto-tuning mechanism TVM and the optimization library cuDNN. Several other methods have been developed, such as the XGBoost \cite{chen2016xgboost} tuner and genetic algorithm (GA), to investigate the possibilities for searching for performance tuning. 

The research done by the authors in \cite{imgClassification} focused on the use of supervised machine learning techniques, namely convolutional neural networks (CNN) and ResNet \cite{he2015deep}, for image classification tasks. The CIFAR10 \cite{krizhevsky2009learning} dataset was utilized as the primary dataset for this investigation. Based on their research, the highest level of accuracy attained was 87.94\% via the use of the CNN model without the incorporation of max-pooling and density function. Nevertheless, the researchers also concluded that it needs a greater amount of processing resources. In \cite{ccalik2018cifar}, the researchers provide a comprehensive analysis of convolutional neural network (CNN) models for image classification tasks using the CIFAR10 dataset. To optimize the applicability of embedded systems, the classification of images is executed within the constraints of restricted storage capacity. By using the architectural framework outlined in their research, the authors successfully achieved a substantial reduction in both training and testing timeframes.

The researchers of the study conducted in \cite{dosovitskiy2020image} investigated the potential utilization of Transformers in addressing the classification of images problem. They succeeded in this by interpreting an image as an ordered set of regions and subjecting it to processing through a conventional transformer encoder. The evaluation of this approach was performed on various datasets, including Imagenet \cite{deng2009imagenet}, CIFAR100 and CIFAR100 \cite{krizhevsky2009learning}, VTAB \cite{zhai2019large}. Experiments were done to compare state-of-the-art models, including ResNet50, ViT, and a hybrid model. The authors have conducted a comparative analysis, evaluating several ViT models including the basic, big, and gigantic variants.

\section{General Architecture of used DNN Models}
This section presents details about the deep learning models used for the implementation in the table below. We applied performance-tuning methods as follows:

\textbf{ResNet50:} 
In ResNet50, residual learning is an approach that incorporates residual blocks, which consist of skip connections enabling the network to circumvent one or more layers. The aforementioned technique effectively mitigates the issue of the vanishing gradient problem often encountered during the training of DNN, hence facilitating the successful training of very complex architectures with a significant number of layers. The fundamental concept revolves on the notion of the network acquiring knowledge of the residual mapping, as opposed to directly conforming to the intended fundamental mapping. This approach facilitates the optimization of the network's weights.

It is a CNN architecture that consists of 50 layers. It is distinguished by its modular design, which includes construction pieces including identity mappings and convolutions. These blocks play a significant role in augmenting the network's capacity to acquire hierarchical characteristics, hence boosting its capability to represent information. It has shown remarkable efficacy in several computer vision applications, owing to its remarkable depth and incorporation of skip connections. Image classification, object identification, and segmentation have consistently demonstrated excellent performance with this architecture. The architectural advancements of this particular model have had a significant impact on following designs of neural networks, establishing it as a fundamental model within the realm of deep learning.

\textbf{Vision Transformer (ViT)}: The ViT significantly transforms the field of image classification by departing from traditional CNN designs. ViT utilizes a transformer-based framework, first developed for problems in natural language processing, and immediately extends it to visual input. The proposed model partitions an input picture into patches that do not overlap, and then applies a linear embedding to every patch. This process generates a series of tokens. The tokens are next subjected to a sequence of self-attention procedures, which empower the model to effectively record comprehensive contextual information over the whole of the picture. The use of the attention mechanism enables effective communication of information across patches, hence enabling the ViT model to comprehend complex patterns and establish connections over great distances within the visual input.

The ViT presents the notion of incorporating positional embeddings to retain the spatial details of the picture patches. By integrating such embeddings alongside the tokenized picture characteristics, the model acquires comprehension of the spatial correlations among distinct areas within the input picture. The transformer encoder is extended with the addition of the classification head, resulting in the generation of the ultimate output. The attention-based method used by ViT obviates the need of manual feature engineering, hence providing a more adaptable and expandable solution for tasks related to image recognition. Despite achieving significant success, the ViT has difficulties in effectively managing spatial hierarchies and encounters limitations in dense prediction tasks as a result of its reliance on patch-based processing. Consequently, researchers are actively engaged in continuing investigations aimed at enhancing and expanding the capabilities of ViT.

\textbf{EfficientNet}: It garnered significant attention in current deep learning research owing to its exceptional ability to effectively manage the trade-off between model size and computing cost. The present model presents an innovative approach to scaling compounds, whereby the network's depth, breadth, and resolution are concurrently optimized to attain exceptional performance under various resource limitations. It effectively uses the compound scaling approach to optimize the utilization of model depth and breadth, hence guaranteeing how the network scales adequately across various computing budgets.

Moreover, it has a revolutionary mobile inverted bottleneck convolution (MBConv) block that intelligently merges inverted residuals alongside linear bottlenecks, hence enhancing the transmission of information and optimizing the utilization of parameters. The inclusion of this unique architectural feature boosts the model's capacity to effectively capture intricate patterns while simultaneously reducing the computing burden.

% Figure ~\ref{fig:efficientnet_arch} shows the approach of the model Scaling in which figure ~\ref{fig:efficientnet_arch}(a) is an example of a baseline network and figure ~\ref{fig:efficientnet_arch}(b) to figure ~\ref{fig:efficientnet_arch}(d) represents conventional scaling that only increases one dimension of network width, depth, or resolution, figure ~\ref{fig:efficientnet_arch}(e) represents the compound scaling method incorporated with EfficientNet that uniformly scales all three dimensions with a fixed ratio\cite{tan2019efficientnet}. In this work, the EfficientNet-B0 variant is used, which is an ImageNet pre-trained model.

\section{Our proposed accelerating training approach}
To alleviate the obstacles arising from CUDA's out-of-memory issues and improve the speed of training execution, we have included many performance-tuning strategies.

\begin{itemize}
    \item \textbf{Gradient accumulation:} The use of gradient accumulation as a strategy enables the simulation of a larger batch size. In the training procedure of a NN, it is customary to divide the training data into smaller subsets referred to as mini-batches, which are then processed in a sequential manner. The neural network generates predictions for batches of data, which may be used to compute the loss with respect to the actual target values. Following this, the reverse pass is performed to compute gradients and modify the weights of the model by shifting them in alignment with these gradients. The Stochastic Gradient Accumulation Equation can be written as:

    \[
\theta_{t+1} = \theta_t - \alpha \left( \frac{1}{M} \sum_{j=1}^{M} \nabla_{\theta} J(\theta; x_{ij}, y_{ij}) \right)
\]
    where \( M \) is the mini-batch size and \( (x_{ij}, y_{ij}) \) represents the \( j \)-th data point in the \( i \)-th mini-batch.
\\
    
\begin{figure}[!htb]
\centering
\includegraphics[width=0.7\textwidth]{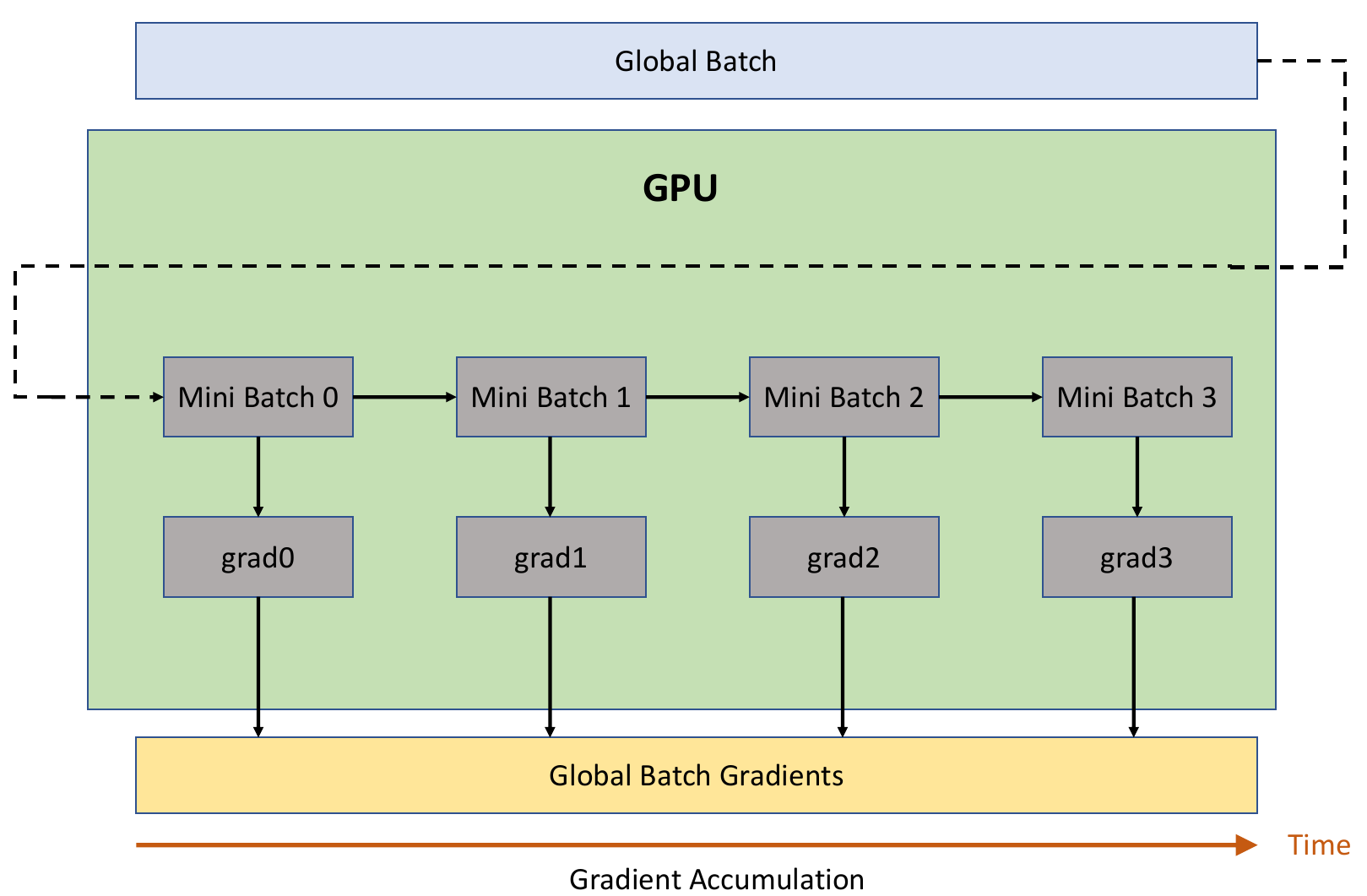}
\caption{\label{fig:grad_accm}Grad accumulation}
\end{figure}

    The method of gradient accumulation modifies the last step of the training process. Instead of doing weight changes after each batch, an alternate strategy entails storing gradient values, progressing to the next batch, and collecting the most current gradients. The weight update is executed sequentially after the completion of many batches by the DNN model.
    
    \item \textbf{Leverage automatic mixed precision\cite{micikevicius2017mixed}:} Typically, deep learning systems often use the training of a neural network using 32-bit floating-point input, denoted as FP32. It is feasible to use a 16-bit floating-point representation for some tasks since the utilization of FP32 entails more time and memory consumption. The use of automated mixed precision and gradient scaling may be achieved via the implementation of functions Autocast and Gradscaler. These functions aim to assign the proper data type to each operation, hence potentially decreasing the computational time and memory usage of the network. This phenomenon often yields the most significant acceleration when the GPU is operating at full capacity. The Mixed Precision Scaling Equation can be written as:

\begin{equation}
S_{\textnormal{mixed}} = \frac{T_{\textnormal{float32}}}{T_{\textnormal{mixed}}}
\end{equation}

  Where:  $S_{\textnormal{mixed}} $ is the speedup achieved with mixed precision; $T_{\textnormal{float32}}$ is the execution time with float32 precision and $T_{\textnormal{mixed}}$ is the execution time with mixed precision.
  
    The mathematical equation~\ref{eq:throughput} represents the computational throughput, equation~\ref{eq:quantization} denotes Quantization Error, and   equation~\ref{eq:dynamicLoss} represents Dynamic Loss Scaling.

    \begin{equation}\label{eq:throughput}
    \textnormal{Throughput} = \frac{N}{T}
    \end{equation}
    Where ``Throughput'' is the computational throughput.
    $N$ is the total number of operations. $T$ is the execution time.
 \begin{equation}\label{eq:quantization}
    \textnormal{Quantization Error} = \frac{1}{N} \sum_{i=1}^{N} \left| x_i - Q(x_i) \right|
\end{equation}

    Where: ``Quantization Error'' is the average error introduced by quantization.
$N$ is the total number of data points, $x_i$ is the original data point and $Q(x_i)$ is the quantized representation of $x_i$.

% The mathematical representation of the Dynamic Loss Scaling equation may be expressed as:
  \begin{equation}\label{eq:dynamicLoss}
    \textnormal{Dynamic Loss Scale} = \textnormal{Base Loss Scale} \times \left( \frac{\textnormal{Old Loss}}{\textnormal{New Loss}} \right)^\beta
\end{equation}
Where: ``Dynamic Loss Scale'' is the dynamically adjusted loss scale, ``Base Loss Scale'' is the initial loss scale, ``Old Loss'' is the loss computed using float32 precision, ``New Loss'' is the loss computed using mixed precision, and $\beta$ is a user-defined constant.

    \item \textbf{Pin Memory:} The pin memory feature is configured as true in the data loader for both the train and test datasets. This setting facilitates the automated allocation of fetched data tensors in memory, resulting in accelerated data transmission to GPUs that support CUDA.
    The mathematical equation~\ref{eq:dynamicMemory} represents the dynamic allocation of memory, equation~\ref{eq:mps} denotes the Memory Pinning Strategy, and equation~\ref{eq:mue} represents the Memory Update Equation.
    \begin{equation}\label{eq:dynamicMemory}
M_t = \arg\max_{M} \sum_{i=1}^{N} \left( \alpha_i \cdot \textnormal{{Sim}}(h_t, m_i) - \beta \cdot P_i \right)
\end{equation}
The above equation represents the dynamic allocation of memory $M_t$ at time step $t$, where $\alpha_i$ is the attention weight, Sim($h_t$,$m_i$) measures the similarity between the current hidden state $h_t$ and memory $m_i$, and $P_i$ denotes the pin status of memory $m_i$.

\begin{figure}[!htb]
\centering
\includegraphics[width=0.7\textwidth]{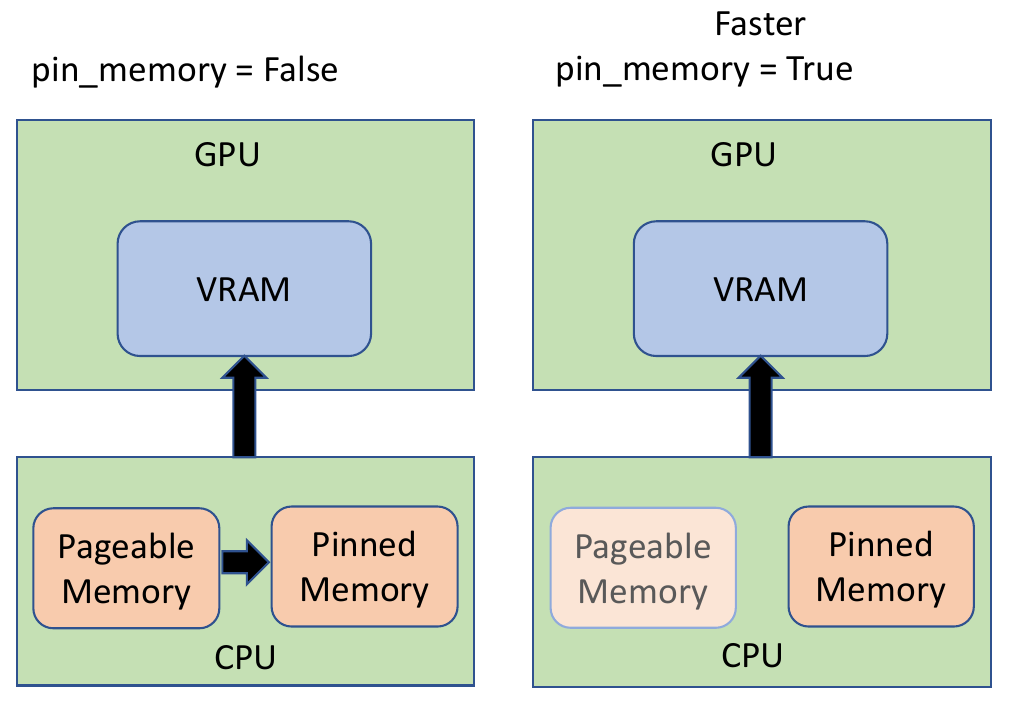}
\caption{\label{fig:pin_mem}Pin memory}
\end{figure}

% Memory Pinning Strategy Equation can be defined as:
\begin{equation}\label{eq:mps}
\textnormal{{Pin\_Score}}(m_i) = \frac{\sum_{t=1}^{T} \left( \gamma \cdot P_i^t - \delta \cdot \left| P_i^t - P_i^{t-1} \right| \right)}{T}
\end{equation}
Where $P^t_i$ is the pin status of memory $m_i$ at time $t$, and $\gamma$ and $delta$ are hyperparameters that control the impact of pinning and changes in pin status over time.

% Memory Update Equation can be defined as:
\begin{equation}\label{eq:mue}
m_i^{t+1} = \rho \cdot m_i^t + (1 - \rho) \cdot \left( \eta \cdot h_t + \xi \cdot P_i^t \cdot \nabla_{m_i} \mathcal{L}(m_i^t) \right)
\end{equation}
Where $\rho$ is the forgetting factor, $\eta$ is the learning rate, and $\nabla_{m_i} \mathcal{L}(m_i^t)$ is the gradient of the loss concerning the memory.
\end{itemize}

% \begin{figure}[!htb]
% \centering
% \includegraphics[width=0.3\textwidth]{resources/auto_precision_scal.png}
% \caption{\label{fig:auto_precision_scal}Auto mixed precision scaling}
% \end{figure}

\section{Datasets}
The CIFAR10 and CIFAR100 datasets were used in our study. The datasets have gained significant recognition within the scientific community for their use in addressing image classification challenges.

\begin{sidewaystable}
  \centering
  \caption{Our studies with performance tweaking yielded the results above.}
  \resizebox{\textheight}{!}{%
    \begin{tabular}{|l|l|l|l|l|l|l|l|l|}
      \hline
      \textbf{Model} & \textbf{Epochs} &
      \textbf{Batch size} & \textbf{Acc \%} & \textbf{F1-score} & \textbf{Exe. time (in hrs)} & \textbf{Train Loss \%} & \textbf{Val loss \%} \\ \hline
      ResNet50 (w/o) & D1:30  & D1:128  & D1:93.07 & D1:93.07 & D1:0.86 & D1:0.11 & D1:0.21   \\ \hline
                     & D2:30 & D2:128 & D2:75.12 & D2:75.12 & D2:0.85 & D2:0.50 & D2:0.95  \\ \hline

      ResNet50 (with) & D1:30 & D1:128 & D1:93.00 & D1:93.00 & D1:0.46 & D1:0.33 & D1:0.26   \\ \hline
                      & D2:30 & D2:128 & D2: 75.23 & D2:75.23 & D2:0.18 & D2:0.14 & D2:0.93   \\ \hline

      ViT (w/o) & D1:10 & D1:70 & D1:97.10 & D1:97.10 & D1:4.2 & D1:0.52 & D1:0.35   \\ \hline
                 & D2:15 & D2:70 & D2:84.30 & D2:84.30 & D2:6.2 & D2:2.75 & D2:2.46  \\ \hline

      ViT (with) & D1:10 & D1:70 & \cellcolor{blue!25}{D1:97} & \cellcolor{blue!25}{D1:97} & \cellcolor{blue!25}{D1:2.0} & D1:0.53 & D1:0.35   \\ \hline
                 & D2:15 & D2:70 & \cellcolor{blue!25}{D2:84.3} & \cellcolor{blue!25}{D2:84.3} & \cellcolor{blue!25}{D2:2.5} & D2:2.6 & D2:2.5  \\ \hline

      EfficientNet (w/o) & D1:20 & D1:96 & D1:98 & D1:98 & D1:1.80 & D1:0.1 & D1:0.02  \\ \hline
                         & D2:20 & D2:96 & D2:90 & D2:90 & D2:1.70 & D2:0.5 & D2:0.18  \\ \hline

      EfficientNet (with) & D1:20 & D1:96 & \cellcolor{green!25}{D1:98} & \cellcolor{green!25}{D1:98} & \cellcolor{green!25}{D1:1.05} & D1:0.12 & D1:0.06   \\ \hline
                         & D2:20 & D2:96 & \cellcolor{green!25}{D2:90} & \cellcolor{green!25}{D2:90} & \cellcolor{green!25}{D2:1.11} & D2:0.5 & D2:0.3  \\ \hline

    \end{tabular}%
  }
  \label{tab:res}
\end{sidewaystable}

\section{Experiments}
The implementation of the project was carried out in Python 3.8, using the Numpy, Matplotlib, PyTorch, and PyTorch Ignite libraries as frameworks for DNNs. The implementation of the EfficientNet model has been carried out using the PyTorch Ignite framework. The performance strategies used include Gradient accumulation, Pin memory, Automatic mixed precision, which aids in increasing the batch size, and the utilization of NVIDIA features such as Autocast and Gradescaler.

In our experimental setup, the selection of hyperparameters for each model was based on its performance and the observed results over several rounds. In this context, a ResNet50 model was constructed and trained from novo, without reliance on a pre-existing model. The ResNet50 model was trained using the following hyperparameters: Adam optimizer, 0.001 is the learning rate, 0.001 for weight decay, a gradient clip of 0.01, and a gradient accumulation of 2.

The ViT model was trained using certain hyperparameters. The learning rate was set to 3e-5 for CIFAR100 and 2e-5 for CIFAR10. The chosen optimizer was Adamw, with a weight decay of 0.01. Additionally, gradient accumulation was used during training, with a value of 15 for CIFAR10 and 20 for CIFAR100.

In the implementation of Efficientnet, the hyperparameters used were as follows: the learning rate was set to 0.01, the optimizer utilized was SGD, the weight-decay value was 1e-3, a dropout rate of 0.2 was applied, and gradient accumulation was performed with a factor of 2.

\section{Results}
In the present section, CIFAR10 will be denoted as D1, whereas CIFAR100 will be referred to as D2. The ResNet50 model achieved an accuracy of 93.07\% (D1 dataset) and an F1-score of 75.12\% (D2 dataset) after 30 epochs. Additionally, the execution time decreased from 0.86 hours to 0.46 hours (D1 dataset) and from 0.85 hours to 0.18 hours (D2 dataset) after performance tweaking.
The ViT model demonstrated high levels of accuracy and F1-score on both the D1 and D2 datasets, achieving 97.1\% and 84.3\% respectively. Furthermore, the execution time of the model was notably decreased by the use of performance-tuning strategies. Specifically, the execution time was lowered from 4.2 to 2 hours for the D1 dataset, and from 6.2 to 2.5 hours for the D2 dataset.
The EfficientNet model has excellent accuracy and F1-score values of 98\% (D1 dataset) and 90\% (D2 dataset), respectively. Additionally, the execution time for the model has been significantly decreased from 1.41 hours to 1.06 hours (D1 dataset) and from 1.42 hours to 1.11 hours (D2 dataset). These results underscore the efficiency of the model. 

Based on an analysis of the aforementioned criteria, it can be concluded that EfficientNet demonstrated a commendable performance in comparison to other models. In Table ~\ref{tab:res}, the term ``Acc'' denotes the accuracy metric of a model. The abbreviation ``w/o'' signifies that the model was trained without any performance tuning, while ``with'' indicates that the model underwent performance tuning during training. The variable ``Exe. time'' represents the execution time of the model, measured in hours.

\begin{figure}[!htb]
\centering
\includegraphics[width=0.8\textwidth, height=3cm]{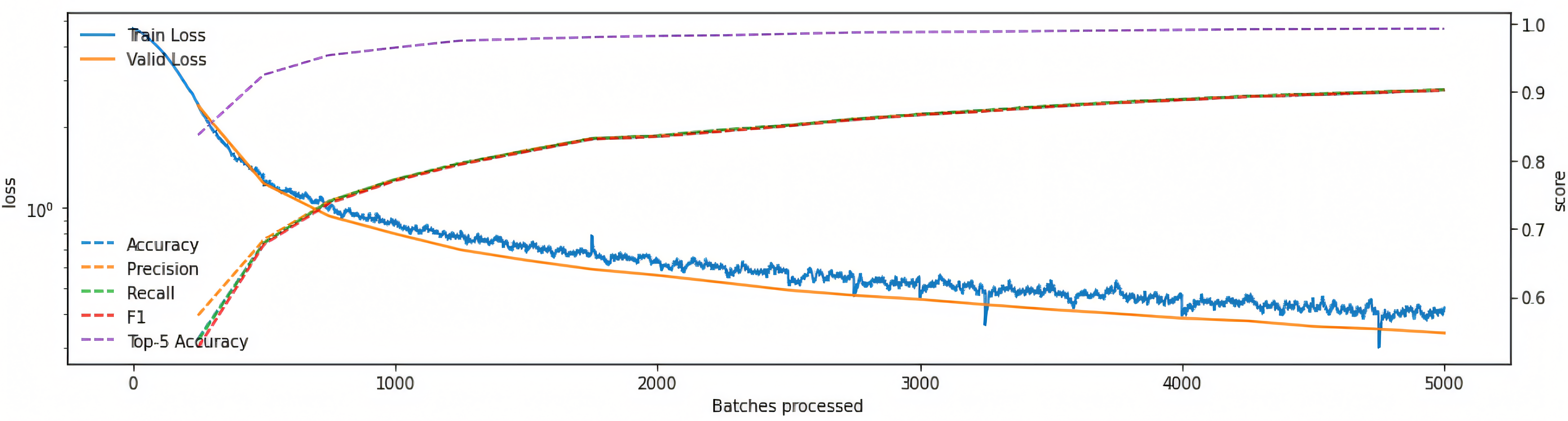}
\caption{\label{fig:res_cifar100} Learning curve for CIFAR100}
\end{figure}

Figure \ref{fig:res_cifar10} and   \ref{fig:res_cifar100} have plotted the results obtained for the EfficientNet model:

\begin{figure}[!htb]
\centering
\includegraphics[width=0.8\textwidth, height=3cm]{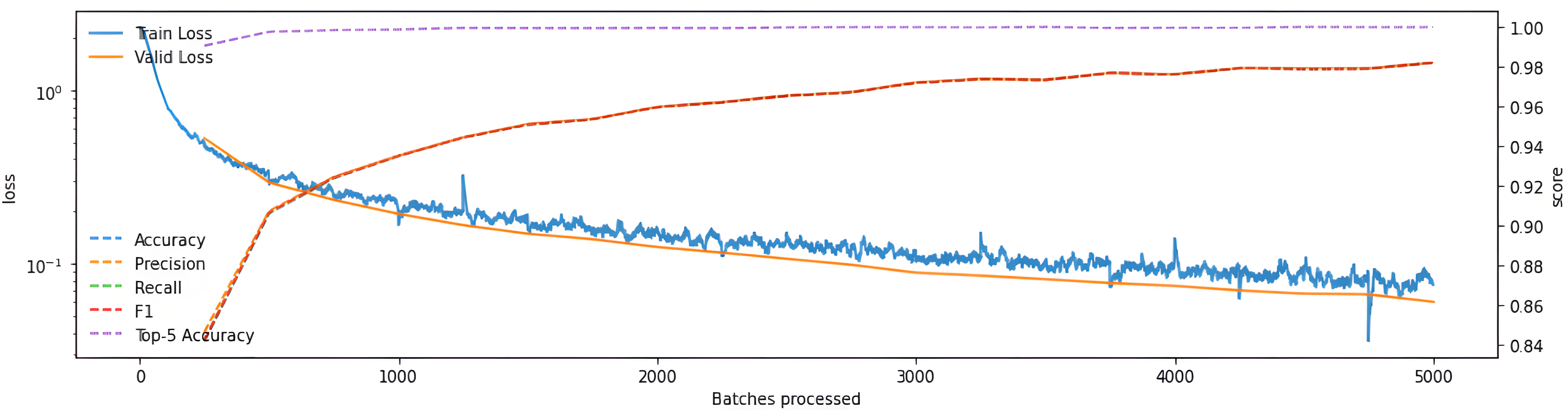}
\caption{\label{fig:res_cifar10} Learning curve for CIFAR10}
\end{figure}

\section{Conclusion and Future Work}
Our research focused on investigating several performance-tuning approaches that might be used to improve the performance of a DNN model. Promising results were reached with the implementation of three state-of-the-art DNNs, notably EfficientNet, and conducting extensive tests on the CIFAR10 and CIFAR100 datasets. The applied techniques included Gradient Accumulation, Pin Memory, and Automatic Mixed Precision. It is worth mentioning that EfficientNet demonstrated superior performance in terms of both accuracy and execution time when compared to other models. The results presented in this study provide significant contributions to the scientific community focused on DNNs.

For future work, we want to investigate other methods for enhancing performance on state-of-the-art models. There is potential for enhancing accuracy via the use of more hardware resources and the exploration of alternative models.

\bibliographystyle{ieeetr}
\bibliography{sample}

\end{document}